%% file: main.tex
\documentclass[runningheads]{llncs}
\PassOptionsToPackage{table}{xcolor}

\usepackage{eccv}

\usepackage{eccvabbrv}
\usepackage{bm}
\usepackage{booktabs} 
\usepackage{tabularx} 
\usepackage{ltablex} 
\usepackage{arydshln}
\usepackage{pifont}
\usepackage{xcolor}
\keepXColumns 

\usepackage[accsupp]{axessibility}  

\newcommand{\cmark}{\textcolor{blue}{\ding{51}}} 
\newcommand{\xmark}{\textcolor{orange}{\ding{55}}} 

\usepackage{dsfont}
\definecolor{colororange}{rgb}{1,0.5,0} 
\definecolor{colorblue}{rgb}{0,0,1}

\usepackage{hyperref}

\usepackage{orcidlink}
\begin{document}

\title{SPIN: Hierarchical Segmentation with Subpart Granularity in Natural Images} 

\titlerunning{SPIN}

\author{Josh Myers-Dean\inst{1} \and
Jarek Reynolds\inst{1}\and
Brian Price\inst{2}\and Yifei Fan\inst{2}\and \\Danna Gurari\inst{1,3}}

\authorrunning{Myers-Dean et al.}

\institute{$~^1$ University of Colorado Boulder, $~^2$ Adobe, $~^3$ University of Texas at Austin}

\maketitle

\input{sections/abstract}

\input{sections/01-intro}
\input{sections/02-related}
\input{sections/03-dataset}
\input{sections/04-evaluation}
\input{sections/05-benchmarking}
\input{sections/06-conclusion}
\paragraph{\bf{Acknowledgments.}}
This work was supported by Adobe Research Gift Funds and utilized the Blanca condo computing resource at the University of Colorado Boulder. Josh Myers-Dean is supported by a NSF GRFP fellowship (\#1917573). We thank the crowdworkers for contributing their time for the construction of SPIN and the authors of our benchmarked models for open-sourcing their work.

\section{Supplementary Materials}
\input{sections/supp}

\clearpage
\bibliographystyle{splncs04}
\bibliography{main}
\end{document}

%% file: sections/abstract.tex
\begin{abstract}
Hierarchical segmentation entails creating segmentations at varying levels of granularity.  We introduce the first hierarchical semantic segmentation dataset with subpart annotations for natural images, which we call SPIN (\textbf{S}ub\textbf{P}art\textbf{I}mage\textbf{N}et).  We also introduce two novel evaluation metrics to evaluate how well algorithms capture spatial and semantic relationships across hierarchical levels.  We benchmark modern models across three different tasks and analyze their strengths and weaknesses across objects, parts, and subparts. To facilitate community-wide progress, we publicly release our dataset at \url{https://joshmyersdean.github.io/spin/index.html}.
\end{abstract}

%% file: sections/01-intro.tex
\section{Introduction}
\label{sec: intro}
When people discuss \emph{hierarchical image analysis} tasks they are typically talking about one of two approaches: \texttt{is-a} relationships~\cite{li2022deep, li2019logic}, which treat categories at different abstraction levels (\eg, a Subaru is a car), or \texttt{is-part-of} relationships, which concentrate on dividing objects into their constituent parts~\cite{de2021part, geng2023gapartnet, hong2021vlgrammar} (\eg, a door is part of a car). While the former has been widely explored, the latter—focusing on object decomposition—has received limited attention in computer vision research. Within segmentation research, the focus of this paper, \texttt{is-part-of} relationships have primarily been explored only for part-whole hierarchies, ignoring finer-grained details, such as subparts (\ie, parts of parts).

Research to subpart-level segmentation granularity is hindered by a scarcity of data.  While synthetic 3D datasets~\cite{de2021part,liu2021cgpart, hong2021vlgrammar, Xiang_2020_SAPIEN} could be used to infer hierarchical segmentations to subpart granularity, it is well-known that models developed with synthetic data typically generalize poorly to natural images (\ie, images taken by a camera)~\cite{Deitke_2020_CVPR}. This lack of annotated natural data has meant that the few models designed to generate subpart granularity segmentations~\cite{wang2024hierarchical, tang2023visual, ding2023visual} could only be evaluated qualitatively on a small number of examples (rather than quantitatively at scale). 

\begin{figure}[t!]
    \centering
    \includegraphics[width=\textwidth]{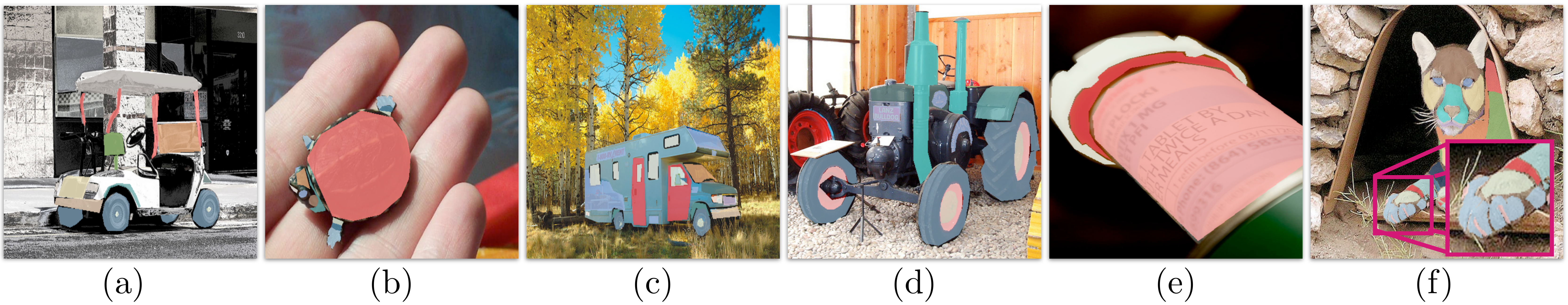}
    \caption{Overview of the diversity of SPIN. Panels (a) and (b) depict subparts unique to specific object class members, such as a roll cage in a car and a shell in a turtle. Panels (c) and (d) illustrate the variability in the number of subparts per object of the same class, with examples of 13 and 6 subparts. Panels (e) and (f) highlight the disparity in image area coverage by different subparts, such as a bottle label (large) versus quadruped claws (tiny).}
    \label{fig: teaser}
\end{figure}

Our work facilitates the development of algorithms for subpart granularity in three key ways. First, we collect over 102,000 subpart segmentations across 203 diverse subpart categories to expand upon PartImageNet~\cite{he2022partimagenet} and create the \textbf{S}ub\textbf{P}art\textbf{I}mage\textbf{N}et (SPIN) dataset—the first to offer subpart annotations for natural images. We release this dataset publicly to foster community progress. Examples of annotated subparts are shown in Fig.~\ref{fig: teaser}.  Second, we introduce two novel metrics to address a shortcoming of the prevailing evaluation methods for hierarchical segmentation, which only assess each granularity level in isolation.  Our newly proposed metrics are intended to work alongside these traditional metrics (\eg, Intersection over Union) and assess how well models capture spatial and semantic relationships across hierarchical levels. Our third key contribution is that we benchmark modern models on SPIN across objects, parts, and subparts, thereby presenting the first comprehensive quantitative evaluation of subpart performance and key areas needed for future improvement.

Advancing hierarchical segmentation to subpart granularity could significantly benefit both research and societal applications. One potential use is in generating more detailed image descriptions, such as for augmented reality experiences, answering visual questions, captioning images, or visual storytelling.  Similarly, this work could be the foundation for individuals to (1) interactively learn how to speak about finer-grained entities for the first time (for children) or in a new language and (2) recall forgotten words, such as when facing temporary or chronic memory injuries or disabilities.  More deeply nested, hierarchical segmentation representations can also enrich tactile visual discovery~\cite{image_assist, image_explorer}, including for use in screen readers employed by visually impaired individuals (\eg, Apple's object-based exploration feature~\cite{apple-image-description}).  Similarly, we believe finer-grained information could facilitate improvements for related problems, including image/video retrieval, image/video editing, automatic magnification, and robotics.

%% file: sections/02-related.tex
\section{Related Works}
\label{sec: related}

\paragraph{\bf{Datasets with Hierarchically Segmented Objects.}} 
Several datasets provide segmentations showing how objects are hierarchically decomposed into their recursively nested parts. A few do so \emph{without semantic labels} specifying the categories of segmented content, including the pioneering Berkeley dataset published in 2001~\cite{martin2001database} and the large-scale SA-1B dataset released in 2023~\cite{Kirillov_2023_ICCV}.  Our work, in contrast, focuses on hierarchically segmenting objects \emph{with category labels}.  Already, numerous datasets semantically decompose objects and their parts in natural images~\cite{zhou2017scene, liang2018look, zhao2018understanding,gong2018instance, wah2011caltech,zheng2018modanet,song2019apollocar3d, jia2020fashionpedia,de2021part,ramanathan2023paco, wei2024ov,chen2014detect,he2022partimagenet} and synthetic images~\cite{chang2015shapenet,mo2019partnet,hong2021ptr,hong2021vlgrammar,liu2021cgpart}.  One dataset, (i.e., ADE20K~\cite{zhou2017scene}) even provides subpart annotations for 10\% of its objects, however its lack of exhaustive annotations for objects and their associated parts~\cite{tang2023visual} impedes it benefit for evaluating prediction models. Extending prior work, we introduce the first exhaustively labelled hierarchical semantic segmentation dataset with subpart annotations for natural images by extending one of the largest part-based datasets~\cite{he2022partimagenet}.  

\paragraph{\bf{Hierarchical Semantic Segmentation Evaluation.}} 
Most evaluation protocols for hierarchical segmentation algorithms assess each hierarchy level \emph{independently}, such as by reporting for object and part categories separately their Intersection over Union scores~\cite{Mo_2019_CVPR, koo2022partglot, deng2020cvxnet, yu2019partnet,liang2018look, wei2024ov, michieli2022edge, wang2015joint, wang2015semantic, tang2023visual, liang2016semantic, tsogkas2015deep}, Average Precision scores~\cite{Sun2023GoingDW, ramanathan2023paco}, and Panoptic Quality scores~\cite{de2021part, wang2024hierarchical, li2022panoptic, li2023panopticpartformer++, tang2023visual}. The one exception is Hierarchical Panoptic Quality~\cite{tang2023visual} which recursively measures the extent to which a predicted hierarchy is complete (e.g., a wheel is a part of a car). Complementing prior work, we introduce evaluation metrics for assessing how well a model captures spatial and semantic hierarchical relationships, such as if a part is perfectly contained in it's whole.

\paragraph{\bf{Hierarchical Semantic Segmentation Algorithms.}} 
Thus far, the focus of methods that predict all entities in a hierarchical decomposition largely has centered on predicting just objects and parts~\cite{He_2023_CVPR, Sun2023GoingDW, wang2015joint}, aligning with existing datasets which only recently increasingly have added part segmentations to object segmentations.  Only a few works, HIPIE~\cite{wang2024hierarchical}, VDT~\cite{ding2023visual}, Semantic-SAM~\cite{li2023semantic}, and ViRReq~\cite{tang2023visual}\footnote{No code is publicly available for VDT, and ViRReq does not offer complete code. At the time of writing, Semantic-SAM has not released their semantic prediction code.}, examined predicting more layers of a hierarchy to also include subparts.  However, these works relied on qualitative assessments for subpart predictions due to a lack of annotated datasets supporting quantitative assessments.  Complementing prior work, we quantitatively evaluate HIPIE and over 20 model variants to show their performance on our new dataset challenge across all three layers of the hierarchy (\ie, object, parts, and  subparts). 

%% file: sections/03-dataset.tex
\section{SPIN Dataset}
We now introduce and characterize our new subpart semantic segmentation dataset that we call SPIN, which is short for \textbf{S}ub\textbf{P}art \textbf{I}mage\textbf{N}et.

\subsection{Dataset Creation}
\label{sec: creation}
\paragraph{\bf{Data Source.}} 
Our dataset extends an existing dataset that provides both object and associated part annotations: PartImageNet~\cite{he2022partimagenet}.  It includes 24,080 natural images across 158 ImageNet categories (112 non-rigid and 38 rigid, \eg animals and vehicles respectively). The authors of PartImageNet used the WordNet~\cite{wordnet} taxonomy to establish these categories and their 11 parent super-categories. The authors then identified 40 part categories pertinent to the super-categories and annotated these in all images.  Every image in PartImageNet contain one semantically segmented object, and we used only those with at least one segmented part. Due to our limited annotation budget, we capped each super-category at 1,200 images. This resulted in the following amounts per super-category: 311 - Aeroplane, 483 - Bottle, 559 - Boat, 634 - Bicycle, and 1,200 each for Biped, Bird, Car, Fish, Quadruped, Reptile, and Snake.

 \paragraph{\bf{Subpart Category Selection.}}
We identified 206 subpart categories to segment for 34 PartImageNet part categories that we identified as being decomposable into subparts.  Of these, 168 subpart categories generalize across many part categories and 38 categories pertain to only a few. For example, in our dataset, \emph{mouths} generalize while \emph{shells} don't, with most reptiles having heads with \emph{mouths} but only one type of reptile having a body with a \emph{shell} (\ie, turtles). We show examples of specific subparts in Fig.~\ref{fig: teaser}(a-b).  A histogram showing how many subparts are assigned to each of the 34 part categories is shown in Fig.~\ref{fig:spin_hist}, and a full list is provided in the supplementary materials.

\begin{figure}[b!] 
  \centering
  \includegraphics[width=\textwidth]{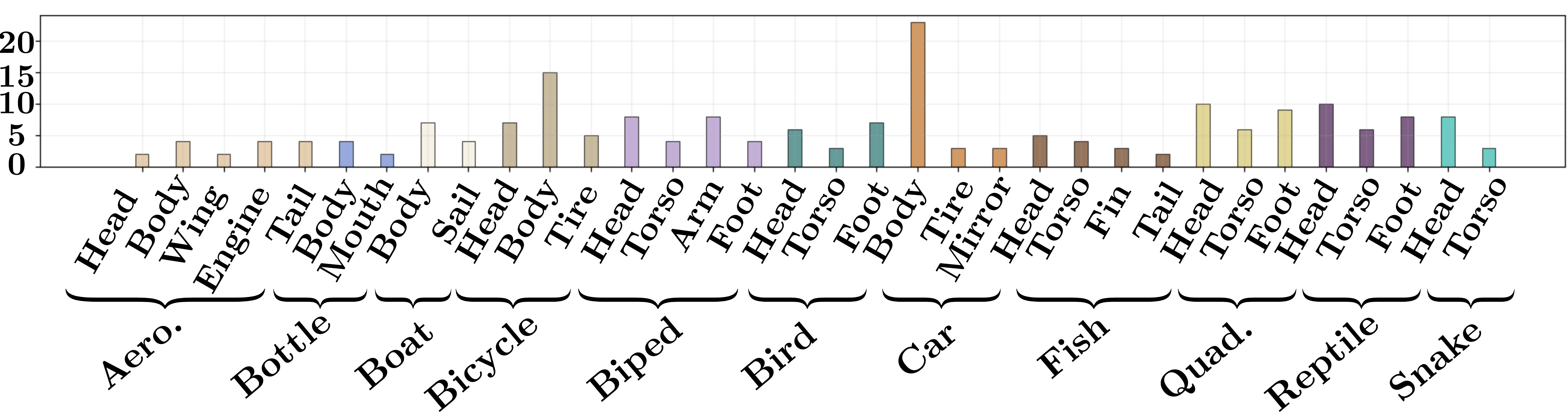}
  \caption{Histogram the number of unique subpart category labels for each of the 34 part categories. (Aero=Aeroplane; Quad=Quadruped)}
  \label{fig:spin_hist}
\end{figure}

We created our subpart taxonomy through a multi-step process.  We identified the general subpart categories by prompting GPT-4~\cite{achiam2023gpt} to list expected subparts for every object-part category in PartImageNet,\footnote{We prompted GPT-4 with "Please list the canonical subparts of a <object>-<part>. Only include subparts that are clearly visible and recognizable to a layperson."} and then three authors edited the list to exclude non-visible subparts (\eg, a skull).  We identified specific subpart categories through manual inspection of at least 15 images per object category, excluding any categories that had ambiguous subpart decompositions; for example, it's not obvious to a lay person what boundaries to use when decomposing a tail into a tip, shaft, and base.  

 \paragraph{\bf{Subpart Annotation Task Design.}}
We created a task interface for collecting all subpart annotations. It presents each image-object pair alongside one of the object's nested parts, outlined with an overlaid polygon. For each part, the task interface has multiple steps. First, users are asked, ``Can you locate the <object>-<part> enclosed in a polygon?" If the target content is present, annotators are then asked for each possible subpart, ``Can you locate any <subpart> on the <object>-<part>?" If the response is yes, annotators segment the subpart. The interface supports segmentation by collecting a series of clicked points to create a connected polygon. It also supports annotating multiple polygons in three scenarios: (1) multiple instances of a subpart (\eg, two eyes), (2) subparts with holes (\eg, coiled snakes), and (3) occlusions breaking a subpart into multiple, disconnected pieces.

 \paragraph{\bf{Annotation Collection.}}
We hired 18 highly trusted annotators from Amazon Mechanical Turk (AMT) who we had vetted through repeated employment for previous segmentation tasks. During data collection, we ensured high annotation quality through five methods: an onboarding qualification test, detailed instructions, live ``office hours" during annotation periods, phased task roll-out for worker feedback, and continuous inspection of submitted results.

 \paragraph{\bf{Dataset Splits.}} Our dataset adheres to the PartImageNet training, validation, and testing splits of 85\%, 5\%, and 10\% of the data, respectively. This results in 8,828 training, 519 validation, and 1,040 test images in our SPIN dataset.

\subsection{Dataset Analysis}
\label{sec:dataset-analysis}
We now characterize SPIN's overall composition as well as its subparts.

 \paragraph{\bf{Overall Dataset Composition.}}
We characterize SPIN with respect to the following seven factors: number of images, number of object categories, total number of annotated objects, number of part categories, total number of annotated parts, number of subpart categories, and total number of annotated subparts.  Results are shown in Table \ref{tab:simple_comp}. As shown, our contribution of 203 entity categories is over a five-order-of-magnitude increase compared to the 40 categories added for PartImageNet~\cite{he2022partimagenet}, while providing nearly an order of magnitude more annotations (\ie 106,324 semantic annotations versus 11,960 part segmentations in PartImageNet).

\begin{table*}[t!]
\centering 
\begin{tabular}{cccccc}
\toprule
\textbf{\#Images} \quad\quad\quad\quad & \textbf{\#Object Cat} \quad\quad\quad\quad & \textbf{\#Part Cat} \quad\quad\quad\quad & \textbf{\#Sub-Part Cat}\\ \midrule
10,387 \quad\quad\quad\quad & 11 (10,387) \quad\quad\quad\quad & 40 (29,818) \quad\quad\quad\quad & 203 (106,324) \\ \bottomrule
\end{tabular}
\caption{SPIN Composition: Each column lists category counts alongside the total number of annotated entities within those categories. (Cat=Categories)}
\label{tab:simple_comp}
\end{table*}

\begin{figure}[t!] 
  \centering
  \includegraphics[width=\textwidth]{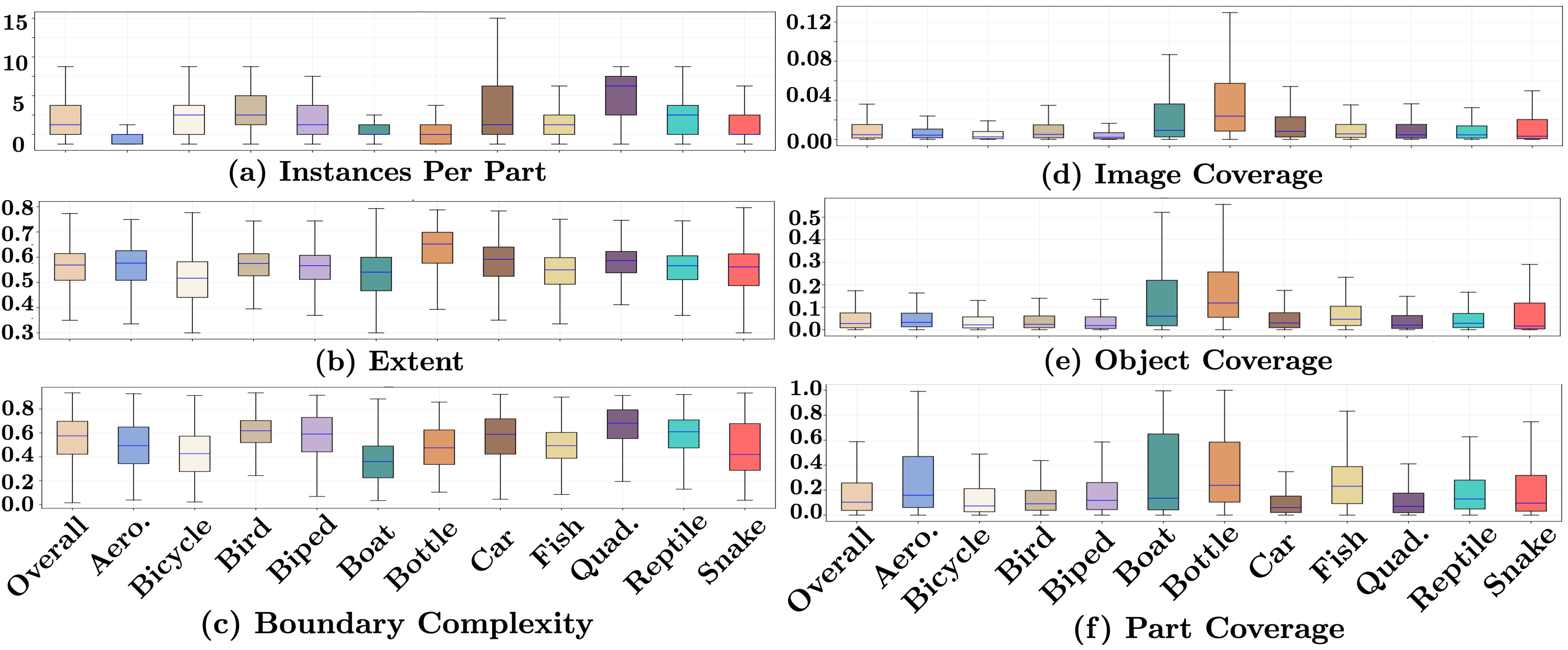}
  \caption{Boxplots showing the distribution of subpart image occupation (left) and boundary complexities per part, per object (right). The blue lines represent medians, bottoms and tops of each box represent the 25th and 75th percentile values respectively, and whiskers represent the most extreme data points not considered outliers. Overall, SPIN's subparts take up a relatively small number of pixels per image, while featuring a range of geometric complexity. (Aero=Aeroplane; Quad=Quadruped)}
  \label{fig:subpart-characteristics}
\end{figure}

 \paragraph{\bf{Subpart Statistics.}} 
We next compute the mean number of subparts per part and characterize the typical appearance of subparts with respect to five metrics:
\begin{itemize}
    \item \textbf{Boundary complexity}: ratio of a subpart’s area to the length of its perimeter (\ie, isoperimetric quotient). Values range from 0 (highly jagged boundary) to 1 (circular).
    \item \textbf{Extent}: the ratio of the area of a contour to the area of its bounding box. Values are in (0,1], where values approaching 0 mean that a contour occupies little area in it's bounding box (\eg, a thin diagonal line) and 1 means that a contour is perfectly contained (\eg, a square).
    \item \textbf{Image coverage}: fraction of image pixels occupied by the subpart. 
    \item \textbf{Object coverage}: fraction of parent object pixels occupied by the subpart. 
    \item \textbf{Part coverage}: fraction of parent part pixels occupied by the subpart. 
\end{itemize}
\noindent
For all six factors, we report both the overall mean as well as the mean with respect to each object category.  Results are shown in Fig.~\ref{fig:subpart-characteristics}. 

Objects tend to have 2 to 5 subparts per part (\ie Fig.~\ref{fig:subpart-characteristics}a; 25th to 75th percentile values), with a total of 4 to 15 subparts. There is diversity across object categories with aeroplanes typically having small variability with only 1 to 2 subparts per part (\ie 25th to 75th percentile range) while car typically exhibits great variability with 2 to 8 subparts per part (\ie 25th to 75th percentile range), as exemplified in Fig.~\ref{fig: teaser}(c-d). We attribute the large range of subpart per part counts to two factors: the intrinsic diversity in cars and the viewing angle. For example, a go-cart viewed from the front will have no windshield, no windows, and two tires, while a bus viewed from a front-right angle will have a windshield, tires, and windows. Such object-specific patterns around the expected prevalence of subparts could be predictive cues for models decomposing detected objects into their recursively nested subparts.

In terms of shape, most subparts' contours occupy the majority of their bounding box (\ie, Fig.~\ref{fig:subpart-characteristics}b, median extent value $\sim$0.7) with moderate boundary complexity (\ie, Fig.~\ref{fig:subpart-characteristics}c, median boundary complexity values around 0.5). Still, we observe high variability for both metrics that we again attribute to the intrinsic diversity in each object category as well as the viewing angle.  For example, a bottle cap or a bicycle tire viewed from the side will be an ellipse versus a nearly perfect circle when viewed from the top (\ie, occupying more of it's bounding box).  Altogether, these findings highlight that our dataset encourages the design of models that can account for a large range of subpart shapes.

Regarding the relative area occupied by each subpart, we find that all subparts occupy small portions of the parent parts, parent objects, and entire image.  For example, across every object category, subparts occupy less than 6\% of the image for at least 75\% of subpart instances (Fig.~\ref{fig:subpart-characteristics}d).  Additionally, the subparts tend to occupy less than 20\% of their parent parts (Fig.~\ref{fig:subpart-characteristics}e; Overall) and less than 5\% of their parent objects (Fig.~\ref{fig:subpart-characteristics}f; Overall).  When adopting size thresholds introduced for the MSCOCO dataset\cite{lin2014microsoft}, where $32^2$ and $96^2$ are thresholds determining whether an object is small, medium, or large, we find 54.10\%  (57,525) of SPIN's subpart annotations qualify as small, 38.08\% (40,488) as medium, and only 7.82\% (8,311) as large.  Most subparts are already considered small according to mainstream research despite that we annotated subparts for a dataset known to typically contain a single, large prominent object in the image. Consequently, embedded in our problem is the well-known challenge of small entity detection~\cite{tong2022deep}, which will only grow as subparts are explored for more complex scenes where less prominent objects must also be hierarchically decomposed.  

When comparing subparts belonging to the rigid object categories (\ie, human-made) versus non-rigid object categories (\ie, animals), we found no clear distinction.  For instance, the mean number of subparts per part (Fig.~\ref{fig:subpart-characteristics}a) is the smallest amount with least variability in aeroplanes and the greatest amount with greatest variability in cars. While prior work has emphasized distinguishing between these two extremes~\cite{he2022partimagenet}, our analysis suggests that rigid and non-rigid objects appear quite similar at the subpart level.

%% file: sections/04-evaluation.tex
\section{Evaluating Hierarchical Consistency}
\label{sec: eval}
We facilitate assessing how well relationships across varying levels of granularity are captured for hierarchical segmentation by introducing two metrics that account for spatial and semantic relationships respectively.

We first introduce the \textbf{Spatial Consistency Score (SpCS)}, which focuses on structural hierarchical alignment rather than semantic alignment.  It indicates the proportion of a child's segmentation (\eg, subparts) that is contained in its super-region's segmentation (\eg, part, object), with values ranging from 0 (no containment) to 1 (perfect containment).  It leverages binarized segmentations distinguishing the foreground entity from the background.  Formally, it is computed as the average containment ratio across all pairs as follows:
\begin{equation}
SpCS = \frac{1}{|\mathcal{R}|} \sum_{(\text{child}, \text{parent}) \in \mathcal{R}}\underbrace{\frac{|\text{child} \cap \text{parent}|}{|\text{child}|}}_{\text{containment ratio}}
\end{equation}
where $\mathcal{R}$ denotes the set of prediction pairs with a ground truth parent-child relationship within a hierarchical structure.\footnote{For a quadruped, for instance, pairs such as $\{$(eyes, head), (chest, torso), (torso, quadruped)$\}$ could be present in $\mathcal{R}$.} $|\mathcal{R}|$ denotes the total number of parent-child pairs, $|\text{child} \cap \text{parent}|$ is the number of pixels in the intersection of the child segmentation with the parent, $|\text{child}|$ is the number of pixels in the child segmentation.

We next introduce the \textbf{Semantic Consistency Score (SeCS)}, which focuses on semantic alignment by measuring how well predictions at different levels (subpart, part, object) logically correspond. For example, an eye (subpart) should correspond to a head (part) and, by extension, to a quadruped (object), rather than illogical associations like a windshield (subpart) with a head (part) or a bottle (object). Let $FG_S$, $FG_P$, and $FG_O$ be the sets of non-background predictions in subpart, part, and object predictions, respectively. Define $X = FG_S \cap FG_P \cap FG_O$ as the intersection of these predictions, representing pixels with consistent foreground labels across levels. For each pixel $x \in X$, with predicted categories $C(S_x)$, $C(P_x)$, and $C(O_x)$ for subparts ($S_x$), parts ($P_x$), and objects ($O_x$), respectively, category entailment is evaluated using $\mathds{1}[C(S_x), C(P_x)]$ and $\mathds{1}[C(P_x), C(O_x)]$, where $\mathds{1}$ is an indicator function returning 1 if the hierarchical entailment is correct, using ground truth relations, otherwise 0.  We compute the entailment measure $M(x)$ as:
\begin{equation}
M(x) = 
\begin{cases}
1 & \text{if } \mathds{1}[C(S_x), C(P_x)] \land \mathds{1}[C(P_x), C(O_x)], \\
0 & \text{otherwise},
\end{cases}
\end{equation}
where $\land$ is the logical \texttt{and} operator. SeCS is the mean of $M(x)$ across all $x \in X$.  Values range from 0 (no semantic coherence) to 1 (perfect semantic coherence). 

%% file: sections/05-benchmarking.tex
\section{Algorithm Benchmarking}
\label{sec: bench}
We now assess modern models' ability to account for hierarchical decompositions to the subpart granularity. 

\subsection{Open-Vocabulary Localization in a Zero-Shot Setting}
\label{sec: seg}
We first evaluate models for open-vocabulary localization in zero-shot mode across three hierarchy levels: object, part, and subpart.  This approach avoids the drawbacks of the alternative options of training models from scratch or fine-tuning, including the computational expense and risks of overfitting to the target task and so diminishing generalizable knowledge.

 \paragraph{\bf{Models.}} 
We benchmark 11 model variants. 

Most related to our work is HIPIE~\cite{wang2024hierarchical}, which is designed for open-vocabulary hierarchical segmentation to the subpart level. It takes as input a list of all candidate categories that can appear and then predicts which categories are present where in the given image. We test two variants with different backbones: ResNet-50~\cite{he2016deep} and ViT-H~\cite{dosovitskiy2020image}. 

We also evaluate 10 variants of foundation models that, by design, are intended for use on a variety of downstream tasks.  Five produce pixel-wise segmentation masks: LISA-7B~\cite{lai2023lisa}, LISA-13B~\cite{lai2023lisa}, GLaMM~\cite{hanoona2023GLaMM}, and PixelLLM-7B~\cite{ren2023pixellm}, and PixelLLM-13B~\cite{ren2023pixellm}. The other five can only produce bounding boxes: CoGVLM~\cite{wang2023cogvlm}, Ferret-7B~\cite{you2023ferret}, Ferret-13B~\cite{you2023ferret}, Shikra~\cite{chen2023shikra}, and Kosmos2~\cite{peng2023kosmos}.  For these models, we analyze their upper bound of their potential performance by notifying the model what object category needs to be located in a given image. We examine two types of requests: ``Can you please locate the <object> in the image?" for objects, and ``Can you please locate the <(sub)part> of the <object> in the image?"\footnote{For subparts, we refer to the parent object rather than the parent part, as broader context aids in processing finer details~\cite{raymond2000psycholinguistics}.} for each part and subpart. We evaluate with these prompts a \textbf{Specific} category by using the original WordNet category from the 158 ImageNet categories as well as a \textbf{General} category by referring to the 11 super-categories.  Consequently, we test each model with four prompts. 

 \paragraph{\bf{Evaluation Metrics.}} 
We use two evaluation metrics for all models: (1) Intersection over Union (IoU) to assess segmentation performance for objects, parts, and subparts independently and (2) our Spatial Consistency Score (SpCS) to assess structural coherence across object, part, and subpart levels. We also use our new Semantic Consistency Score (SeCS) to assess semantic coherence between predicted hierarchical levels for the only relevant model, HIPIE (it predicts from a list of candidate categories rather than for the target, single category).

\begin{table}[!t]
\centering 
\resizebox{\textwidth}{!}{%
\begin{tabular}{lcccccccccccccc}
\toprule
 & &\multicolumn{2}{c}{\bf mIoU$_{\bm{S}}$} & \multicolumn{2}{c}{\bf mIoU$_{\bm{P}}$} & \multicolumn{2}{c}{\bf mIoU$_{\bm{O}}$} & \multicolumn{2}{c}{\bf SpCS - Avg} & \multicolumn{2}{c}{\bf SpCS - S2P} & \multicolumn{2}{c}{\bf SpCS - P2O} \\
\cmidrule(r){3-4} \cmidrule(r){5-6} \cmidrule(r){7-8}  \cmidrule(r){9-10} \cmidrule(r){11-12} \cmidrule(r){13-14}
\bf Method& \bf  \# Params. & \bf Specific & \bf General & \bf Specific & \bf General & \bf Specific &\bf General & \bf Specific & \bf General& \bf Specific & \bf General& \bf Specific & \bf General&\\
\midrule
HIPIE~\cite{wang2024hierarchical} (R-50)& 200M & 0.80 & 0.10 & 8.05 & 7.22 & 51.36 & 13.87 & 97.82 & 98.77 & 100 & 100 & 95.64 & 97.54  \\
HIPIE~\cite{wang2024hierarchical} (ViT-H) & 800M & 0.90 & 0.92 & 7.21 & 8.23 & 66.77 & 21.69 & 98.04 & 99.25 & 100 & 98.50 & 96.08 &96.06  \\ 
PixelLLM~\cite{ren2023pixellm} & 7B &  9.08 & 9.24 & 32.37 & 31.37 & 83.93 & 79.87 & 73.79 & 82.46 & 69.57 & 79.20 & 89.49 & 90.73 \\
PixelLLM~\cite{ren2023pixellm} & 13B& 9.53 & 10.13 & 32.04 & 32.96 & 82.90 & 79.96 & 84.86 & 83.06 & 82.13 & 80.87 & 92.60 & 88.90 \\
LISA~\cite{lai2023lisa} & 7B & 9.61 & 9.14 & 27.72 & 27.28 & 86.23 & 83.36 & 79.30 & 77.44 & 74.61 & 72.93 & 92.10 & 89.78 \\
LISA~\cite{lai2023lisa}  & 13B & \bf 11.52 & \bf11.55 & 32.29 & 31.36 & \bf 87.78 & 85.45 & 86.28 & 79.98 & 82.87 & 75.70 & \bf 96.02 & 92.41 \\
GLaMM~\cite{hanoona2023GLaMM} & 7B & 11.03 & 11.00 & \bf 39.41 & \bf{40.00} & 86.29 & \bf{86.31} & \bf 87.12 & \bf{87.75} & \bf 84.21 & \bf{84.96} & 95.72 & \bf{95.38} \\ 
GLaMM - FT & 7B & 24.25 & 24.56 & 59.37 & 60.76 & 86.42 & 91.08 &87.13 & 87.95 & 75.93 & 85.16 & 90.23 & 96.04 \\
\hdashline
Ferret~\cite{you2023ferret} & 7B & 7.22 & 7.30 & 25.69 & 26.54 & 47.84 & \bf 47.98 & 78.70 & 73.02 & 46.50 & 65.68 & 93.65 & \bf{94.60} \\
Ferret~\cite{you2023ferret} & 13B & 6.37 & 6.25 & 23.67 & 23.65 & \bf 47.99 & 47.50 & 81.04 & 77.30 & 48.70 & 70.98 & 94.17 & 95.40 \\
CoGVLM~\cite{wang2023cogvlm} & 17B & \bf 13.94 & \bf 14.29 & \bf 38.13 & \bf 38.56 & 46.47 & 43.65 & 76.33 & 77.20 & 72.37 & 73.32 & 86.19 & 87.59 \\
Shikra~\cite{chen2023shikra} & 7B & 8.41 & 8.72 & 26.50 & 27.42 & 45.50 & 29.14 & 68.57 & 67.01 & 63.82 & 66.76 & 81.85 & 67.89 \\
Kosmos2~\cite{peng2023kosmos} & 2.6B & 3.45 & 3.59 & 19.00 & 19.48 & 48.63 & 48.89 & \bf 82.74 & \bf{81.60} & \bf 78.34 & \bf{77.03} & \bf 96.01 & 95.58 \\ \hdashline
SAM~\cite{Kirillov_2023_ICCV} & 630M & 49.61 & 49.61 & 69.23& 69.23 & 90.06 & 90.06& 86.42 & 86.42 & 83.85 & 83.85 & 92.30 & 92.30 \\
SAM + CoGVLM & 630M & 19.88 & 19.98 & 50.93 & 50.90 & 77.94 &75.09& 66.71 & 67.33 & 80.50 & 80.73 & 60.65 & 61.44 \\ \bottomrule
\end{tabular}}
\caption{Performance of modern models with respect to localization at three granularity levels---subparts (mIoU$_S$), parts (mIoU$_P$), objects (mIoU$_O$)---and hierarchical consistency at two levels---subparts to parts (SpCS-S2P), parts to objects (SpCS-P2O), and their mean (SpCS - Avg). Results are shown for hierarchical semantic segmentation models (described in Section~\ref{sec: seg}), hierarchical object detection models (Section~\ref{sec: seg}), and interactive segmentation models that perform category-agnostic localization (Section~\ref{sec: vf_loc}).  The method families are separated by dashed lines in that order. Best performing scores are shown in bold. } 
\label{tab: loc_results}
\end{table}

 \paragraph{\bf{Localization Results (mIoU).}} 
Results are shown in Table~\ref{tab: loc_results}.  

We observe a consistent trend across all models: localization performance is best for object segmentation, followed by increasingly worse performance for more granular categories of parts and then subparts.  The best score for localizing objects is 86 followed by 40 for parts and 14 for subparts.  The worst scores come from the only model directly designed for hierarchical segmentation, HIPIE.  It does worse across all granularity levels than the foundation models; \eg, for subpart localization, 1 vs 3 from the worst-performing foundation model, Kosmos2.  We attribute the poor performance to HIPIE's limited training data (\eg, limited part vocabulary), which only included part-level categories from Pascal Parts~\cite{chen2014detect}.  Moreover, HIPIE's heuristic grouping assumes that parts are the sum of their subparts, which is not the case in SPIN (\eg, unlabelled ``space" exists between a cheek an eye in a head).

We explored to what extent entity size is correlated to the resulting IoU scores by using linear regression for the better-performing foundation models.  Due to space constraints, we provide further experimental details and results in the supplementary materials.  In summary though, our findings suggest that size has little correlation to performance for objects but greater correlations for more granular levels of parts and subparts. We suspect that one contributing factor is the sensitivity of the IoU metric for entities occupying a small number of pixels, as small errors are especially prone to yield dramatic changes in the IoU scores for such cases.

We next explored the impact of a prompt's category specificity on model performance by comparing performance for when general versus specific terms are employed. Most models perform worse when trying to locate \emph{objects} categorized under general terms (\eg, `quadruped') rather than specific ones (\eg, `dog'). For example, CoGVLM performs 2.82 percentage points (pp) worse with general categories compared to specific categories. Similarly, the 13B variant of LISA has a 2.33pp performance drop and its 7B counterpart has a 2.87pp performance drop. The exceptions to this are GLaMM, Ferret, and Kosmos2, whose performance are relatively stable and unaffected by the level of category specificity. A potential reason is that these models utilize training data that explicitly use both general and specific terms. For example, GCG~\cite{hanoona2023GLaMM}, a source of training data for GLaMM, refers to entities with both abstract and specific terms (\eg, person and toddler). Overall, this variance highlights the challenge in choosing an appropriate level of category specificity for \emph{objects}.  However, the story changes when trying to locate \emph{parts} and \emph{subparts}, with the performance gap between general and specific categories falling to within 1pp.  An interesting direction for future work is better understanding how and why an entity's granularity level influences models' abilities to overcome abstraction challenges.

We also examine how the training data of the Large Language Models (LLM) Llama~\cite{touvron2023llama}, since it serves as the basis for the majority of foundation models evaluated in our study.  Specifically, we analyze the empirical frequency (\ie, total occurrence) of uni-grams (\eg, eye) and n-grams (\eg, eye of the quadruped) within the dataset using the $\infty$-gram API~\cite{liu2024infini}, which uses byte array pattern matching to find all occurrences of a given BPE tokenized~\cite{sennrich2015neural} $n$-gram. Due to space constraints, results are provided in the supplementary materials.  In summary though, we observe a decline in average uni-gram occurrence frequency when increasing granularity (\eg, part to subpart). An interesting direction for future work is to remedy this imbalance, such as by designing parameter-efficient methods to restructure the text embedding space of LLMs to recognize hierarchical relationships, akin to MERU~\cite{desai2023hyperbolic} restructuring CLIP~\cite{radford2021learning}-space for image-level \texttt{is-a} relationships.

 \paragraph{\bf{Performance When Training on Our SPIN Dataset.}}
We next examine the potential benefit of our SPIN training data for modern models.  To do so, we fine-tuned the top-performing GLaMM algorithm~\cite{hanoona2023GLaMM} for all hierarchy levels in SPIN, and we refer to this variant as GLaMM-FT.  Results are shown in Table~\ref{tab: loc_results}.  

Overall, we observe a considerable performance boost.  For example, when prompting with general categories, we observe a 13.56 percentage point (pp) increase (123\% relative increase) for subparts, a 20.76pp increase (51.90\% relative increase) for parts, and a 4.77pp increase for objects (5.51\% relative increase).  While such performance gains are promising, we still observe very low absolute scores for subparts, indicating their challenge for modern models.  We suspect a promising direction for future work is providing cost-effective ways to further increase the amount of available training data, such as through synthetic data creation and 2D projections from 3D models.

 \paragraph{\bf{Hierarchical Consistency Results.}}
SpCS results are shown in Table~\ref{tab: loc_results}.  Similar trends are observed for part-object and subpart-part relationships.  Part-object pairs consistently exhibit high SpCS-P2O scores, indicating models tend to correctly predict parts within their corresponding whole objects.  This provides evidence that models' have some understanding of spatial containment among part-object relationships.  A potential reason for this is that models are indirectly trained on part-whole relationships, because they are already taught that an object is a part of a scene.   The aforementioned trend is slightly worse when examining consistency between subparts and parts (\ie, SpCS-S2P), with scores varying from 46.50 (\ie, Ferret 7B) to 84.21 (\ie, GLaMM) when employing general terms for objects and from 65.68 (\ie, Ferret 7B) to 84.96 (\ie, GLaMM) when employing more specific categories for objects. This variability, coupled with low mIoU scores for subparts, reveals that models often predict subparts outside their intended part boundaries, underscoring poor performance in properly contextualizing subparts within an object.  

\subsection{Interactive Segmentation for Vocabulary-Free Localization}
\label{sec: vf_loc}
We next assess to what extent models can \emph{localize} entities at all hierarchy levels in our dataset if they are notified where to look in an image.  This sets an upper bound of what is possible. To do so, we benchmark the state-of-the-art interactive segmentation model, SAM~\cite{Kirillov_2023_ICCV}, which takes as input a coarse marking on an image specifying where to create the segmentation.  We use bounding boxes as our coarse image markings, because they are shown to outperform alternatives~\cite{Myers-Dean_2024_WACV} when using SAM.  We explore the upper bound performance for this model by using the ground truth object, part, and subpart segmentations to generate the bounding boxes (\ie, by using $[x_{min}, y_{min}, x_{max}, y_{max}]$ for each entity's pixels).  Results are shown in Table~\ref{tab: loc_results}. 

Despite the idealized guidance from the bounding box prompts, SAM's performance is still imperfect.  As granularity increases, localization performance consistently decreases by about 20pp when going down a granularity level (\eg, objects to parts), revealing that SAM struggles more as segmentation requests become more fine-grained.  Still, our findings demonstrate SAM's potential for yielding considerably higher quality hierarchical segmentations than observed from existing localization models benchmarked in the previous section.  Compared to the top-performing zero-shot segmentation-based model (\ie, GLaMM), SAM achieves a boost of 38.61pp for subparts, 20.23pp for parts, and 3.75pp for objects, alongside similar SpCS scores. 

We also evaluate a fully-automated system with SAM, by incorporating bounding box predictions from the top-performing bounding box producing model, CoGVLM.  Results are shown in Table~\ref{tab: loc_results}.  We similarly observe that it outperforms the leading zero-shot segmentation method, GLaMM, in segmenting \emph{subparts} and \emph{parts} (9.24 percentage point (pp) boost for subparts and 10.87pp for parts).  We attribute these performance gains to SAM's ability to leverage a location prior (\eg, bounding box), whereas models like GLaMM rely solely on language priors. In contrast, this approach performs worse than the leading zero-shot segmentation method, GLaMM, for \emph{objects}, with a 11.22pp decrease.  We attribute this performance drop to CoGVLM's inferior performance in consistently and accurately detecting bounding boxes. 

\subsection{Recognizing Hierarchical Semantics in a Zero-Shot Setting} 
\label{sec: recognition}
We finally evaluate the ability of models to \emph{recognize} hierarchical semantic labels for objects in images (\ie, without localization). We assess this in an idealized scenario in which we notify models what to look for as well as where to look in the image.  This sets an upper bound of what may be possible with such models as they are explicitly told \emph{where} to look.

 \paragraph{\bf{Models.}}
We evaluate eight off-the-shelf model variants: CoGVLM~\cite{wang2023cogvlm}, Ferret-7B~\cite{you2023ferret}, Ferret-13B~\cite{you2023ferret}, Shikra~\cite{chen2023shikra}, Osprey~\cite{Osprey}, Kosmos2~\cite{peng2023kosmos}, ViP-Llava-7B~\cite{cai2023vipllava}, and ViP-Llava-13B~\cite{cai2023vipllava}. All prompts to the models resemble the following basic request: \texttt{"Is there a <object> in the <region>?"} for objects and \texttt{"Is there a <(sub)part> of the <object> in the <region>?"} for parts and subparts.  Regions are specified either with segmentations (ViP-Llava, Ferret, Osprey) or bounding boxes (CoGVLM, Shikra, Kosmos2). We generate these regions directly from the ground truth annotations (\eg, bounding boxes as described in Section~\ref{sec: vf_loc}) for objects, parts, and subparts.  We prompt models for each category known to be in the region, one at a time. An answer is `yes' if `yes' is present in the model's output, and otherwise `no'. As done in Section~\ref{sec: seg}, we test both general and specific categories in the prompts.  

 \paragraph{\bf{Evaluation Metrics.}} 
We use \emph{accuracy} to evaluate at each granularity level. 

\begin{table}[!t]
    \centering
  \resizebox{\linewidth}{!}{%
    \begin{tabular}{lcc cc cc cc}
    \toprule
    \bf Method & \bf \# Params. & \bf Prompt & \bf mACC$_{\bm S}$ & \bf mACC$_{\bm{SS}}$ & \bf mACC$_{\bm P}$ & \bf mACC$_{\bm{PS}}$ & \bf mACC$_{\bm O}$ & \bf mACC$_{\bm{OS}}$ \\
\midrule
Shikra~\cite{chen2023shikra} & 7B & Bbox & 66.12 & 61.28 & 79.67 & 74.56 & 77.04 & 82.05 \\
Ferret~\cite{you2023ferret} & 7B& Bbox& 59.01 & 57.19 & 59.95 & 66.17 & 86.78 & 89.21 \\
Ferret~\cite{you2023ferret} & 13B & Bbox & 50.31 & 69.42 & 61.72 & 67.84 & 88.93 & 89.02 \\ \hdashline
CoGVLM~\cite{wang2023cogvlm} & 17B & CoT & 42.16 & 45.54 & 81.75 & 83.31 & 87.15 & 83.67 \\
Kosmos2~\cite{peng2023kosmos} & 2.6B & SG & 79.12 & 64.32 & 74.23 & 68.53 & 84.55 & 78.78 \\ \hdashline
Osprey~\cite{Osprey} & 7B & Mask & 77.43 & 55.95 & 86.30 & 81.20 & 96.43 & 84.85 \\
ViP-Llava~\cite{cai2023vipllava} & 7B & Mask &53.57 & 58.52 & 64.13 & 64.96 & 71.62 & 65.25 \\
ViP-Llava~\cite{cai2023vipllava} & 13B & Mask &94.78 & 97.43 & 99.90 & 99.80 & 98.33 & 95.06 \\
    \bottomrule
    \end{tabular}%
  }
  \captionof{table}{Accuracy metrics for object hierarchical decomposition: general (mACC$_S$, mACC$_P$, mACC$_O$) and specific (mACC$_{SS}$, mACC$_{PS}$, mACC$_{OS}$) categories across subparts, parts, and objects. (CoT = Chain of Thought; SG = Self-Grounding)}
  \label{tab: acc}
\end{table}

 \paragraph{\bf{Results.}} 
Results are reported in Table~\ref{tab: acc}. 

While most models struggle overall to recognize categories in images (despite being told where to look in the images), one model has nearly perfect accuracy: ViP-Llava-13B.  One hypothesis for this atypically strong performance is the model lacks understanding and instead always answers yes when prompted about the presence of categories.  We explored this with further experiments using adversarial examples that are instead expected to always answer ``no''.\footnote{Results are shown in the supplementary materials for two prompts asking ``Is the category not present'' and ``Is the [different category] present''.}  However, our findings from these experiments are none-conclusive. We observe the model can answer ``no'', but it also suffers in this case from a significant decrease in accuracy for the adversarial prompts. This performance drop could be due to a separate issue that the model struggles with negation (\ie, the reversal curse~\cite{berglund2023reversal}). Altogether, this evaluation represents the best-case scenario for performance; it is likely to fall further in real-world applications due to imperfect human inputs or model-generated segmentations, both of which we have shown to exhibit poorer performance at finer granularity levels, as discussed in Sections~\ref{sec: seg} and~\ref{sec: vf_loc}.

We observed mixed outcomes regarding which granularity levels are most difficult to recognize.  For Ferret-13B and Osprey, an increase in granularity correlates with improved accuracy; \eg, transitioning from subparts to objects using general category names results in a 28.62pp increase for Ferret-13B and a 19.00pp increase for Osprey. Conversely, for most other models, object identification consistently outperforms subpart identification. The differential performance across granularity levels may partly stem from the disparate prevalence of tokens in the training data across different granularity levels, particularly within the Llama dataset (as previously discussed). For parts and subparts, the mixed results observed—where parts sometimes outperform subparts (\eg, CoGVLM) and vice versa (\eg, Ferret-7B)—could be attributed to their relative frequency in the training data. Future research could focus on examining how foundation models differentiate responses based on prompt granularity and on enriching datasets like GRIT~\cite{you2023ferret} with hierarchical relationships to improve model understanding of intra-entity relationships.  Additionally, future work could explore enhancing hierarchical-relationship understanding so success at different hierarchical levels can facilitate finding entities at other granularity levels.

We also observed mixed outcomes when comparing the efficacy of specific versus general categories. Half the models performed better using general terms for subparts, with a similar pattern observed in parts and objects. A potential reason for this discrepancy, compared to the relatively consistent results in Sec~\ref{sec: seg}, is that localization models often rely on feature similarities, such as CLIP~\cite{radford2021learning} features. Even if a model does not capture every nuance of a category query, it still attempts find a match based on overall similarity to known categories (\eg, mini-van vs. van) and is less likely to reject user queries~\cite{wu2023see}.  Conversely, interactive understanding models, designed to maximize accuracy, may strictly search for precise category matches within their trained representations. Given our task's structure (\ie, answers are restricted to `yes' or `no'), these models are predisposed to outright reject the query (\ie, answer `no') if an exact match is absent, rather than proposing close or related category matches. 

%% file: sections/06-conclusion.tex
\section{Conclusion} 
We introduce SPIN, the first dataset challenge for hierarchical segmentation at the subpart granularity in natural images.  Our analysis reveals SPIN's characteristics, our two new evaluation metrics enable gauging algorithmic proficiency in capturing the spatial and semantic relationships across hierarchy levels, and our benchmarking across three tasks demonstrates that models struggle to recognize and segment subparts in SPIN, even under idealized guidance. We publicly release SPIN to encourage further advancements in hierarchical segmentation, sharing the dataset at \url{https://joshmyersdean.github.io/spin/index.html}.

%% file: sections/supp.tex
\noindent
This document supplements the main paper with the following:
\begin{enumerate}
    \item SPIN dataset creation. (supplements \textbf{Section 3.1})
    \item Crowdsourcing implementation. (supplements \textbf{Section 3.1)}
    \item SPIN analysis. (supplements \textbf{Section 3.2})
    \item Benchmarking models' implementations. (supplements \textbf{Section 5})
    \item Analysis of model performance based on region size vs. IoU. (supplements \textbf{Section 5.1})
    \item Analysis of granularity uni/n-gram frequency in Llama training data. (supplements \textbf{section 5.1})
    \item Adversarial prompting experiments for ViP-Llava 13B. (supplements \textbf{Section 5.1})
    \item Qualitative results from benchmarked models. (supplements \textbf{Section 5.1)}
\end{enumerate}

\section{SPIN Dataset Creation}

\subsection{Candidate Subpart Taxonomy}
To identify candidate subpart categories for each object-part pair in PartImageNet~\cite{he2022partimagenet}, we prompted GPT-4~\cite{achiam2023gpt} with ``Please list the canonical subparts of a <object>-<part>. Only include subparts that are clearly visible and recognizable to a layperson.''  The results were the following:

\begin{itemize}
\item \textbf{Quadruped-Head:} ears, eyes, nose, mouth, tongue, teeth, whiskers, forehead, cheeks, chin
\item  \textbf{Quadruped-Torso:} shoulders, back, belly, chest, ribs
\item \textbf{Quadruped-Foot (leg):} hip, thigh, knee, shin, ankle, foot, toes, claws, pads, hoof
\item  \textbf{Quadruped-Tail:} base, midsection, tip
\item  \textbf{Biped-Head:} ears, eyes, nose, mouth, tongue, teeth, cheeks, forehead, chin, hair
\item \textbf{Biped-Torso:} shoulders, chest, back, abdomen, waist, hips
\item \textbf{Biped-Arm (includes hand):}  shoulder, upper arm, elbow, forearm, wrist, hand, fingers, thumb
\item \textbf{Biped-Foot (includes leg):} hip, thigh, knee, calf, ankle, foot, toes
\item  \textbf{Biped-Tail:} base, midsection, tip
\item \textbf{Fish-Head:} eyes, mouth, gills, nostrils
\item \textbf{Fish-Torso:} scales, lateral line, dorsal surface, ventral surface
\item \textbf{Fish-Fin:} rays, spines, lobes, base
\item \textbf{Fish-Tail:} caudal peduncle, caudal fin, upper lobe, lower lobe
\item \textbf{Bird-Head:} beak, eyes, nostrils, ears, crown, nape
\item \textbf{Bird-Torso:} chest, belly, back, flanks
\item \textbf{Bird-Wing:} primaries, secondaries, coverts, alula
\item \textbf{Bird-Foot (includes leg):} thighs, knees, shanks, toes, talons
\item \textbf{Bird-Tail:} rectrices, pygostyle
\item \textbf{Snake-Head:} eyes, mouth, nostrils, fangs, tongue
\item \textbf{Snake-Torso:} scales, ventral plates, dorsal surface
\item \textbf{Reptile-Head:} eyes, mouth, nostrils, tongue, teeth, ears
\item \textbf{Reptile-Torso:} scales, belly, back, sides
\item \textbf{Reptile-Foot (includes leg):} thigh, knee, ankle, toes, claws
\item \textbf{Reptile-Tail:} base, midsection, tip
\item \textbf{Car-Body:} hood, trunk, roof, doors, windows, fenders, bumpers
\item \textbf{Car-Tire (includes all of the car wheel):} tread, sidewall, bead, rim, hubcap, valve stem
\item \textbf{Car-Side-Mirror:} mirror glass, housing, adjustment mechanism
\item \textbf{Bicycle-Head:} handlebars, stem, fork, front brake
\item \textbf{Bicycle-Body:} frame, chain, pedals, crankset, gears
\item \textbf{Bicycle-Seat:} saddle, seat post, clamp
\item \textbf{Bicycle-Tire (includes all of the wheel):} tread, sidewall, tube, rim, spokes, hub
\item \textbf{Boat-Body:} hull, deck, keel, rudder, bow, stern
\item \textbf{Boat-Sail:} mainsail, jib, boom, mast, rigging
\item \textbf{Aeroplane-Head:} cockpit, nose, windshield, radome
\item \textbf{Aeroplane-Body:} fuselage, cabin, cargo hold, doors, windows
\item \textbf{Aeroplane-Wing:} flaps, ailerons, slats, wingtips
\item \textbf{Aeroplane-Tail:} vertical stabilizer, horizontal stabilizer, rudder, elevators
\item \textbf{Aeroplane-Engine:} turbine, fan blades, exhaust, nacelle
\item \textbf{Bottle-Body:} main chamber, label, base
\item \textbf{Bottle-Mouth:} opening, neck, lip, cap
\end{itemize}

\subsection{Final Subpart Taxonomy}
As described in the main paper, we manually edited the results from GPT-4 to finalize the taxonomy.  The final resulting taxonomy is as follows (parent objects listed in bold, followed by each part and its associated subparts): 

\begin{itemize}
\item{\bf{Aeroplane}} $\rightarrow$ Head: nosecone and windshield. Body: windows, doors, windshield, and decals. Wing: body and flaps. Engine: intake, outer casing, propeller, and cap. Tail: rudder, vertical stabilizer, horizontal stabilizer, and decals. 

\item{\bf{Bottle}} $\rightarrow$ Body: label, shoulder, base, and neck. Bottle-mouth: rim and cap. 

\item{\bf{Boat}} $\rightarrow$ Body: cockpit, deck, hull, bowsprit, decals, pontoon, and window. Sail: vertical beam, horizontal beam, decals, and sail. 

\item{\bf{Bicycle}} $\rightarrow$ Head: handlebars, brake levers, headlight, bell or horn, grips, mirror, and tassel. Body: seat tube, top tube, down tube, head tube, fork, chainring, pedals, cranks, suspension, foot rest, stem, fender, axle, light, and parental control handle. Tire: tire, rim, spokes, fork and hub. 

\item{\bf{Biped}} $\rightarrow$ Head: eyes, ears, nose, mouth, teeth, forehead, jaw, and neck. Torso: chest, abdomen, back, and shoulders. Arm: forearm, elbow, upper arm, wrist, palm, dorsal area, fingers, and shoulders. Foot: toes, heel, sole, and dorsal area. 

\item{\bf{Bird}} $\rightarrow$ Head: eyes, beak, nostrils, forehead, neck, and cheek. Torso: breast, back, and belly. Foot: toes, claws, shank/forearm, thigh, knee, webbing, and ankle. 

\item{\bf{Car}} $\rightarrow$ Body: door, window, roof, hood, trunk, bumper, decal, light, siren, grille, fender, windshield, windshield wiper, license plate, spoiler, exhaust, roll cage, ladder, plow, seat, hopper, trailer, and spare wheel. Tire: rim, tire, and hub cap. Side-mirror: mirror glass, housing, and mount. 

\item{\bf{Fish}} $\rightarrow$ Head: eyes, mouth, gills, snout, and neck. Torso: neck, dorsal surface, ventral surface, and side. Fin: dorsal fins, pectoral fins, and ventral fins. Tail: lower lobe and upper lobe

\item{\bf{Quadruped}} $\rightarrow$ Head: eyes, ears, nose, mouth, horns, tusk, forehead, cheek, neck, and snout. Torso: back, chest, belly, side, shoulders, and neck. Foot: toes/hoof, claws, pads, dorsal area, heel, shank/forearm, knee/elbow, thigh/upper arm, and wrist/ankle. 

\item{\bf{Reptile}} $\rightarrow$ Head: eyes, mouth, nostrils, tongue, neck, forehead, ears, casque, hood, and throat pouch. Torso: shell, belly, side, back, neck and dorsal fin. Foot: toes, webbing, pads, shank/forearm, knee, thigh/upper arm, wrist/ankle, and fin. 

\item{\bf{Snake}} $\rightarrow$ Head: eyes, mouth, horn, nostrils, tongue, hood, forehead, and cheek. Torso: belly, back, and rattler.
\end{itemize}

\subsection{PartImageNet Filtering}
We removed the 29 images from PartImageNet with only the background class annotated (\ie, no part annotations) because they couldn't support subpart annotation.  We also excluded the following six PartImageNet's part classes that have ambiguous subpart decompositions: biped tails, bird tails, quadruped tails, bird wings, and bicycle seats. 

Next, we restricted every PartImageNet category to include at most 1,200 images by using stratified sampling to preserve PartImageNet's original train, validation, and test split distribution. When sampling, we prioritized images containing the most parts from the part taxonomy to enhance the amount and diversity of annotated subpart annotations. 

\section{Crowdsourcing}
\subsection{Annotation Tool}
\begin{figure}[t!] 
  \centering
  \includegraphics[width=24em]{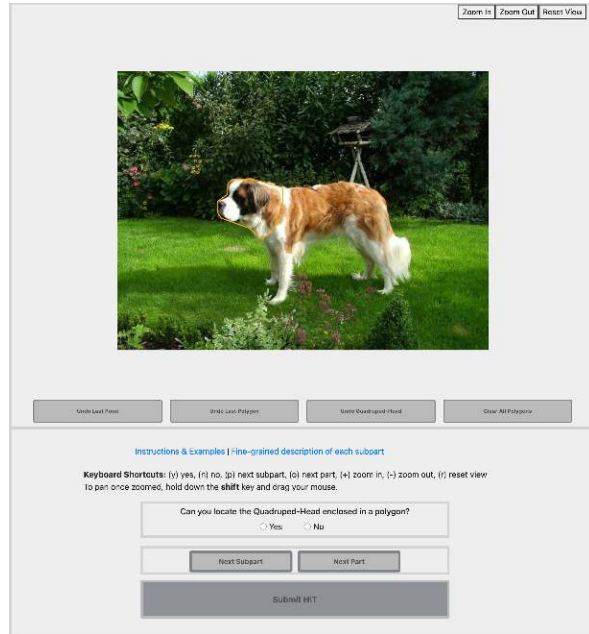}
  \caption{Interface AMT crowdworkers used to create SPIN's ground truth annotations.}
  \label{fig:spin_interface}
\end{figure}

Fig.~\ref{fig:spin_interface} provides a screenshot of our crowdsourcing interface. We included in its design zooming functionality to enable more precise boundary annotations for subparts occupying tiny portions of images. 

\subsection{Crowdsourcing Implementation}
We encouraged high-quality results in multiple ways.  First, every annotator had to complete an initial onboarding task by passing a qualification test with five challenging annotation scenarios. Afterward, we provided a link to a 25-page PPT presentation that provided both generic annotation instructions (matching closely what they already used for their previous object-part annotation task with our team) as well as task-specific instructions clarifying for each super-category, textually and visually, how to annotate each subpart.  These can be found at \url{https://joshmyersdean.github.io/spin/index.html}. After releasing tasks to AMT, we kept a live dialogue channel open with all annotators both by answering questions through email as well as via regular open Zoom sessions that individuals could join to solicit input. To further control quality, we released tasks to AMT in a phased rollout where we released all tasks for a single super-category (\eg, Quadruped, Bicycle) in a series of small batches before moving to the next super-category so workers could sharpen and retain skills on each category before moving to the next one. Following the completion of initial mini-batches per super-category, we manually spot-checked the results for potential worker confusion and provided individual feedback as needed until we found no further concerns. Additionally, throughout the annotation process, we manually inspected suspicious results, such as when workers flagged many parts and subparts as not being present, had missing subpart segmentations, or were outliers in the amount of time they took to complete tasks. We replaced unsuitable annotations as needed in addition to two authors inspecting every annotation and performing corrections as needed.

Toward's providing equitable compensation, we based HIT reward amounts on the maximum number of subparts a worker could encounter when annotating a particular super category.  This design choice addressed the issue that there is high variation in the number of possible subparts per object category.  To determine the pay amount, we conducted in-house testing to find the mean task duration relative to each super category. We found that paying 10 cents per subpart resulted in compensation above the United State's federal minimum wage.  This rate resulted in compensating workers \$1.10 per image for less complex categories like Boat, which only featured eleven potential subparts, versus \$2.80 or \$2.90 per image for more complex categories like Bicycles and Cars, respectively.

\section{SPIN Analysis}

\subsection{Prevalence of Subpart Annotations per Part Category}
\begin{figure}[t] 
  \centering
  \includegraphics[width=\textwidth]{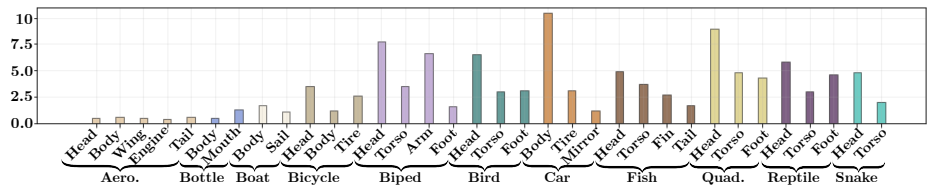}
  \caption{Histogram visualizing the number of subpart-part category occurrences (in the thousands) across the SPIN dataset spanning each of the 34 part categories. We note that the biped and quadruped head, and the car body feature the most significant number of subpart occurrences within their parent part. (Aero=Aeroplane; Quad=Quadruped)}
  \label{fig:spin_hist}
\end{figure}

We next characterize the subparts we augmented to the dataset by computing the frequency of subpart annotations per part category across SPIN's 11 supercategories with results shown in Fig.~\ref{fig:spin_hist}. 

We also characterize the subparts we augmented to the dataset by computing the frequency of subpart annotations per part category across SPIN's 11 supercategories, with results shown in Fig.~\ref{fig:spin_hist}. We observe that car bodies exhibit the most subpart annotations per part category. We attribute this finding to the fact that the car body part category features the highest concentration of subpart categories (23 subparts) relative to all other part categories in SPIN. Additionally, the subpart categories within the car body part category, such as door, window, bumper, decal, and lights, often require multiple annotations per subpart. We observe similar trends in quadruped, biped, and reptile heads. Although this part category features fewer subparts than car bodies, they each contain subpart categories that often require multiple annotations to entirely segment, such as eyes, ears, nostrils, and cheeks. We also find that many of SPIN's images relative to these particular supercategories are biased toward these specific parts as they are often the principal area of focus in the image. For example, a reptile's feet could feature 20 toes and claws, yet a reptile's feet are unlikely to be the focus of an image. Last, these part categories also belong to super categories featuring 1200 images. In contrast, supercategories like aeroplane, bottle, and boat feature 311, 483, and 559 images, naturally lending them fewer subpart annotations than bipeds, quadrupeds, and cars. 

\subsection{Presence of Holes in Subparts}
We evaluate the presence of holes within individual subparts in SPIN. For each subpart, we count how many holes it contains, defined by a polygon embedded within another. 

Overall, we observe a relatively low presence of holes within subparts, with only 2.86\% of subparts containing holes. Cars have the largest proportion of subparts containing holes at 13.54\% and bottles have the lowest number of holes at 0.11\%. Intuitively, a car contains subparts that naturally have holes, such as tires (which rims and hubcaps reside within), as well as grilles (which license plates and headlights reside within). In total, 6/11 object categories contain subparts in which greater than 1\% contain holes: Aeroplane (3.58\%), Bicycle (2.76\%), Boat (6.85\%), Car (13.54\%), Reptile (1.19\%), and Snake (4.70\%). Of the remaining 5 object categories, all contain subparts with less than 1\% having holes: Biped (0.56\%), Bird (0.64\%), Bottle (0.11\%), Fish (0.12\%), and Quadruped (0.54\%). 

Among \emph{all subpart instances containing holes}, all have an average of less than 2.  Boat has the highest average at 1.79 holes, and Fish has the lowest at 1.00 holes. A contributing reason for a scarcity of holes within subparts is that subparts are the finest level of granularity within an object, and thus other subparts typically do not reside within a subpart to create a hole.

\subsection{Multiple Polygons in Subparts}
\begin{figure}[t] 
  \centering
  \includegraphics[width=\textwidth]{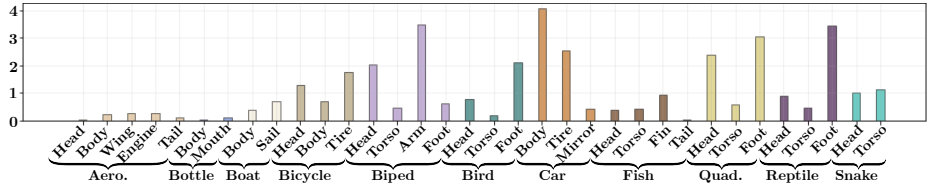}
  \caption{Histogram visualizing the number of subparts (in the thousands) that required multiple polygons to annotate spanning each of the 34 part categories. We note that biped arms, quadruped and reptile feet, and car bodies feature the most subpart occurrences requiring multiple polygons to annotate. (Aero=Aeroplane; Quad=Quadruped)}
  \label{fig:spin_hist_multi_subparts}
\end{figure}

The prevalence of requiring multiple polygons per subpart is shown with respect to objects in Fig.~\ref{fig:spin_hist_multi_subparts}. 

We find that 31.21\% (33,188) of subpart annotations have more than one polygon.  In other words, subpart categories belonging to these object categories contain multiple polygons in the semantic annotations.  Most subpart occurrences requiring multiple polygons occur for biped arms, quadruped and reptile feet, and car bodies.  We attribute this finding to the intrinsic properties of these particular subparts and the viewing angle. For instance, the biped arms, reptile, and quadruped feet often exhibit 5-10 fingers and toes, and reptile and quadruped feet sometimes feature claws that can require an additional 5-10 polygons. In addition, car bodies can contain 2-20 windows depending on the vehicle type, as well as 2 lights and 4 tires, underscoring why this category has the most significant number of multi-polygon subpart annotations of all object categories.

\begin{figure}[t!] 
  \centering
  \includegraphics[width=\textwidth]{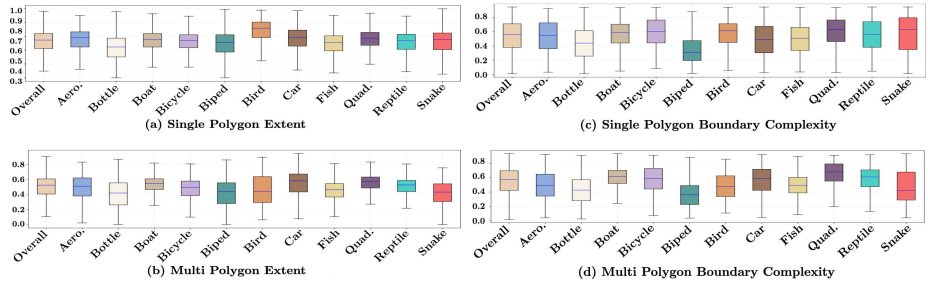}
  \caption{Subpart extent and boundary complexity relative to the number of polygons required to segment the subpart, grouped by their respective super categories.}
  \label{fig:spin_ecc_bc_bp}
\end{figure}

We next characterize subparts consisting of multiple polygons based on two metrics: 1) \emph{Extent}: the ratio of a subpart's area to it's bounding box. Values are in (0,1], where values approaching 0 mean that a contour occupies little area in it’s bounding box (\eg, a thin diagonal line) and 1 means that a contour is perfectly contained (\eg, a square).; and \emph{Boundary complexity}: ratio of a subpart's area to the length of its perimeter (\ie, isoperimetric quotient). Values range from 0 (highly jagged boundary) to 1 (circular). For regions consisting of multiple polygons, we record the mean of each metric for each polygon. We compute the average of each metric across all constituent polygons in a subpart's annotation. Results are shown in Figure \ref{fig:spin_ecc_bc_bp}.

Regarding shape in single—and multi-polygon subpart annotations, the primary trend we observe is that single-polygon annotations take up the majority of their bounding boxy. In contrast, multi-polygon annotations tend to only occupy ~50\% of their bounding box (\ie, Fig.~\ref{fig:spin_ecc_bc_bp} a, b, values closer to 1 compared to b). Intuitively, single polygons may take up more space as there is less background captured in the bounding box (\ie, there are less overall background pixels).

We also see a similar trend in boundary complexity, especially in Bicycles, as their respective inter-quartile ranges get much wider and further away from 0.5 in multi-polygon part annotations compared to single-polygon subpart annotations, ultimately exhibiting moderate albeit more complex boundary complexity among multi-polygon subpart annotation versus single-polygon subpart annotations (\ie, Fig.~\ref{fig:spin_ecc_bc_bp} c, d). We see this trend in bicycles more than in cars because subparts like tires on a bicycle occupy a more significant portion of the bicycle's area compared to a tire on a vehicle, which occupies much less area relative to the object. 

\section{Model Benchmarking}
\subsection{Design of Benchmarked Models}
For each benchmarked model, we report the number of parameters, visual encoder, LLM (\ie, text encoder), capabilities, and model source for inference in Table~\ref{tab: models}. For SAM~\cite{Kirillov_2023_ICCV}, we adopt the commonly used ViT-H~\cite{dosovitskiy2020image} variant. For models producing bounding boxes, we post-process predicted object detections for each semantic category by converting them into a single pixel-wise mask to create a semantic segmentation.  

We also report the specific prompts used for each model, as they vary with each model's official implementation. For objects, <region> is the name of the object (\eg, quadruped, antelope) and for (sub)parts, <region> is the name of the (sub)part and the object (\eg, eyes of the quadruped, eyes of the antelope). We use the same prompts for models that have both 7B and 13B variants (Ferret, LISA, PixelLLM, ViP-Llava).

\paragraph{\bf{Open-Vocabulary Localization Prompts.}}
\begin{itemize}
    \item \textbf{Ferret:} ``Please locate the <region> in this image. Only locate the <part> but locate all instances of the <part>.'' We omit the second sentence when doing object-level localization. 
    \item \textbf{CoGVLM:} ``Please describe the <object> in detail and provide its coordinates [[x0, y0, x1, y1]].''
    \item \textbf{Shikra:} ``Can you point out <region> in the image <image> and provide the coordinates of its location?'' Where <image> is the tokenized image.
    \item \textbf{Kosmos2:} ``<grounding><phrase> the <region> </phrase>''
    \item \textbf{LISA:} ``Please segment the <region> in this image.''
    \item \textbf{GLaMM:} ``Please segment the <region> in this image.''
    \item \textbf{PixelLLM:} ``Please segment the <region> in this image.''
\end{itemize}

\paragraph{\bf{Interactive Understanding Prompts.}}
\begin{itemize}
    \item \textbf{Kosmos2:} ``<phrase>Is there a <region> in the image? Think step-by-step.''
    \item \textbf{Ferret:} ``Is this <mask><pos> a <region>? Only answer yes or no with no other output.'' Where <mask><pos> is the tokenized mask with positional encoding.
    \item \textbf{Osprey:} ``Is this <mask><pos> a <region>? Only answer yes or no with no other output.'' Where <mask><pos> is the tokenized mask with positional encoding.
    \item \textbf{Ferret:} ``Is this <mask><pos> a <region>? Only answer yes or no with no other output.'' Where <mask><pos> is the tokenized mask with positional encoding.
    \item \textbf{ViP-Llava:} ``Is there a <region> in the blue region? Answer yes or no.'' Where ``blue region'' is the overlayed ground truth segmentation mask of the region.
    \item \textbf{Shikra:} ``For this image <image>, I want a simple and direct yes or no answer to my question: Is there a <region> in this region <boxes>?'' in which <image> is the tokenized image, and <boxes> is the ground truth bounding box.
\end{itemize}

\begin{table}[!t]
\centering 
\resizebox{\textwidth}{!}{%
\begin{tabular}{@{}lcccccr@{}}
\toprule
\bf{Model}         & \bf{Parameters}   &\bf{Visual Encoder} & \bf{LLM} & \bf{Open-Vocab Localization} & \bf{Interactive Understanding} & \bf{Model Source} \\ \midrule
HIPIE~\cite{wang2024hierarchical}                   & 200M                  & ResNet-50~\cite{he2016deep}         & BERT~\cite{devlin2018bert}      & \cmark                            & \xmark      & github.com/berkeley-hipie/HIPIE                        \\
HIPIE~\cite{wang2024hierarchical}                   & 800M                   & ViT-H~\cite{dosovitskiy2020image}          & BERT~\cite{devlin2018bert}         & \cmark                            & \xmark        & github.com/berkeley-hipie/HIPIE                       \\
LISA~\cite{lai2023lisa}                   & 7B                   & SAM ViT-H~\cite{Kirillov_2023_ICCV}          & LLaVa-7B-v1-1~\cite{liu2023llava}         & \cmark                            & \xmark          &  github.com/dvlab-research/LISA                    \\
LISA~\cite{lai2023lisa}                   & 13B                   & SAM ViT-H~\cite{Kirillov_2023_ICCV}           & Llama-2-7B~\cite{touvron2023llama}         & \cmark                            & \xmark         & github.com/dvlab-research/LISA                     \\
GLaMM~\cite{hanoona2023GLaMM}                & 7B                       & SAM ViT-H~\cite{Kirillov_2023_ICCV}              & Vicuna-7B~\cite{vicuna2023}         & \cmark                          & \xmark         & github.com/mbzuai-oryx/groundingLMM                      \\
PixelLLM~\cite{ren2023pixellm}              & 7B                      & CLIP-ViT-L/14~\cite{radford2021learning}               & LlaVA-7B~\cite{liu2023llava}         & \cmark                           & \xmark     & github.com/MaverickRen/PixelLM                          \\
PixelLLM~\cite{ren2023pixellm}              & 13B                      & CLIP-ViT-L/14~\cite{radford2021learning}               & LlaVA-llama-13B~\cite{liu2023llava,touvron2023llama}         & \cmark                           & \xmark           & github.com/MaverickRen/PixelLM                    \\
CoGVLM~\cite{wang2023cogvlm}                 & 17B                         & EVA2-CLIP-E~\cite{fang2023eva}           & Vicuna1.5-7B~\cite{vicuna2023}         & \cmark                           & \cmark         &  github.com/THUDM/CogVLM                      \\
Ferret~\cite{you2023ferret}                 & 7B                      & CLIP-ViT-L/14~\cite{radford2021learning}              & Vicuna-7B~\cite{vicuna2023}         & \cmark                        & \cmark        & github.com/apple/ml-ferret                       \\
Ferret~\cite{you2023ferret}                 & 13B                      & CLIP-ViT-L/14~\cite{radford2021learning}              & Vicuna-13B~\cite{vicuna2023}         & \cmark                        & \cmark     & github.com/apple/ml-ferret                          \\
Shikra~\cite{chen2023shikra}                 & 7B                   & CLIP-ViT-L/14~\cite{radford2021learning}              & Vicuna-7B~\cite{vicuna2023}          & \cmark                            & \cmark       & github.com/shikras/shikra                        \\
Kosmos2~\cite{peng2023kosmos}                & 1.6B                      & Unspecified               & MAGNETO Transformer~\cite{wang2023magneto}         & \cmark                           & \cmark     & huggingface.co/docs/transformers/en/model\_doc/kosmos-2                          \\
ViP-Llava~\cite{cai2023vipllava}              & 7B                       & CLIP-ViT-L/14~\cite{radford2021learning}                  & Vicuna-7B~\cite{vicuna2023}          & \cmark                              & \cmark    & huggingface.co/docs/transformers/main/en/model\_doc/vipllava     \\
ViP-Llava~\cite{cai2023vipllava}              & 13B                       & CLIP-ViT-L/14~\cite{radford2021learning}                  & Vicuna-13B~\cite{vicuna2023}          & \cmark                              & \cmark    & huggingface.co/docs/transformers/main/en/model\_doc/vipllava     \\ 
Osprey~\cite{Osprey}                 & 7B                        & CLIP-ConvNeXt-L~\cite{radford2021learning}              & Vicuna-7B~\cite{vicuna2023}          & \xmark                              & \cmark    & github.com/CircleRadon/Osprey    \\ \bottomrule
\end{tabular}
}
\caption{Overview of benchmarked foundation models with respect to their parameters, encoder types, LLM, task capabilities, and model source. (B=billions; M=millions).}
\label{tab: models}
\end{table}

\subsection{HIPIE Analysis}
Despite poor localization from HIPIE, it is worth noting that HIPIE has interesting hierarchical performance results.  First, it achieved nearly perfect spatial consistency between parts and objects (\ie, SpCS-P2O) and perfect spatial consistency between subparts and parts (\ie, SpCS-S2P).  In other words, when HIPIE predicted parts are always perfectly contained within their parent parts which, in turn, are typically perfectly contained within their parent objects. When examining HIPIE's semantic consistency with SeCS metrics for \emph{general} and \emph{specific} categories, we find ResNet-50 outperforms ViT-H for general categories (85.85\% vs. 73.38\% SeCS) despite ViT-H's higher object mIoU. This suggests that ViT-H's increased computational power does not enhance part/subpart accuracy, but rather only object-level performance. ViT-H also shows a higher abstention rate from subpart predictions (\ie, does not predict segmentations) than ResNet-50 (35.77\% vs. 24.71\%). For specific categories, both backbones score high on SeCS (94.58\% for ResNet-50 and 100\% for ViT-H) but abstain 84\% of the time, likely due to the large, similar \emph{specific} category list (154 specific vs. 11 general categories), highlighting issues like differentiating `box turtle' from `mud turtle' in specific categories.  In contrast, all labels in the general categories share little similarity.

\begin{table}[!h]
\centering
\begin{tabular}{lcccccc} 
\toprule
 & \multicolumn{2}{c}{\bf Object} & \multicolumn{2}{c}{\bf Part} & \multicolumn{2}{c}{\bf Subpart} \\
\cmidrule(lr){2-3} \cmidrule(lr){4-5} \cmidrule(lr){6-7}
\bf Model & \( \bm{R^2} \) & \bf p-value & \( \bm{R^2} \) & \bf p-value & \( \bm{R^2} \) & \bf p-value \\ 
\midrule
HIPIE R50 & 0.00 & \cellcolor{orange!50} & 0.00 & \cellcolor{orange!50}& 0.00 & \cellcolor{orange!50}\\
HIPIE ViT-H & 0.00 & \cellcolor{orange!50} & 0.00 & \cellcolor{orange!50}& 0.00 & \cellcolor{orange!50}\\
PixelLLM 7B~\cite{ren2023pixellm} & 0.00 & \cellcolor{orange!50} & 0.67 & \cellcolor{blue!25} & 0.44 & \cellcolor{blue!25} \\
PixelLLM 13B~\cite{ren2023pixellm} & 0.00 & \cellcolor{orange!50} & 0.67 & \cellcolor{blue!25}& 0.37 & \cellcolor{blue!25}\\
LISA 7B~\cite{lai2023lisa} & 0.04 & \cellcolor{orange!50} & 0.63 & \cellcolor{blue!25} & 0.28 & \cellcolor{blue!25} \\
LISA 13B~\cite{lai2023lisa} & 0.01 & \cellcolor{orange!50} & 0.62 & \cellcolor{blue!25}& 0.49 & \cellcolor{blue!25}\\
GLaMM~\cite{hanoona2023GLaMM} & 0.01 & \cellcolor{orange!50} & 0.55 & \cellcolor{blue!25}& 0.35 & \cellcolor{blue!25}\\ \hdashline
Ferret 7B~\cite{you2023ferret} & 0.11 & \cellcolor{blue!25} & 0.64 & \cellcolor{blue!25}& 0.28 & \cellcolor{blue!25}\\
Ferret 13B~\cite{you2023ferret} & 0.04 & \cellcolor{orange!50} & 0.65 & \cellcolor{blue!25} & 0.35 & \cellcolor{blue!25}\\
CoGVLM~\cite{wang2023cogvlm} & 0.05 & \cellcolor{orange!50}& 0.01 & \cellcolor{orange!50}& 0.10 & \cellcolor{blue!25}\\
Shikra~\cite{chen2023shikra} & 0.07 & \cellcolor{blue!25}& 0.04 & \cellcolor{orange!50}& 0.20 & \cellcolor{blue!25}\\
Kosmos2~\cite{peng2023kosmos} & 0.07 & \cellcolor{blue!25}& 0.76 & \cellcolor{blue!25}& 0.36 & \cellcolor{blue!25}\\ 
\bottomrule
\end{tabular}
\caption{Impact of size on predicting IoU for open-vocabulary localization models. We report Pearson $R^2$ coefficients and $p$-values. Blue cells represent statistically significant results for $\hat{\beta_1}$ in $\text{IoU}\sim \hat{\beta}_1\log(\text{region size}) + \hat{\beta}_0$ ($p < 0.001$), and orange represents results that are not statistically significant. Above the dashed line represents segmentation models, and below represents models that output bounding boxes.}
\label{tab: reg}
\end{table}

\section{Analysis of Region Size vs. IoU}
To examine the influence that region size has on segmentation results, we ran a linear regression, $\text{IoU}\sim \hat{\beta}_1\log(\text{region size}) + \hat{\beta}_0$, and calculated the Pearson $R^2$ correlation coefficients for each model at every granularity level, also noting the median $p$-value of $\hat{\beta}_1$ to assess the significance of region size on IoU performance. Results are shown in Table~\ref{tab: reg}. We include HIPIE in the table but exclude it in our discussion as its poor results skew trends. 

Overall, we observe mixed outcomes. No significant positive correlation is observed for objects, with Pearson correlation $R^2$ values between $0.003$ and $0.105$ (median $p$-value $\approx 0.009$), suggesting that an object's segmentation size does not strongly predict IoU scores. Conversely, a positive correlation is noted for parts, indicated by $R^2$ values ranging from $0.014$ to $0.759$ (median $p$-value $\approx 1\mathrm{e}{-10}$), implying that larger parts may correspond to higher IoU scores, depending on the model. Subparts show a weaker positive correlation, with $R^2$ values from $0.102$ to $0.491$ (median $p$-value $\approx  2\mathrm{e}{-20}$), highlighting that while segmentation size impacts performance, it is not the predominant factor.

\section{Analysis of granularity uni/n-gram frequency in Llama training data}
\begin{figure}[!h]
    \centering
    \includegraphics[height=11em,width=\textwidth]{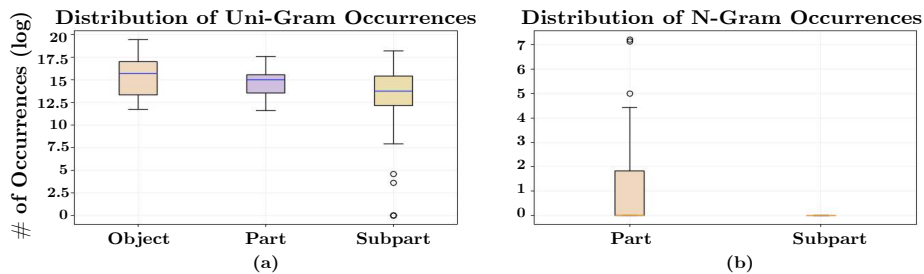}
    \caption{(a) Distribution of uni-gram (\eg, quadruped, head, eyes) across the RedPajama dataset for objects, parts, and subparts. (b) Distribution of n-gram (\eg, head of the quadruped, eyes of the quadruped) across the RedPajama dataset for parts and subparts. We show a log scale to account for wide-range values.}
    \label{fig: token_dist}
\end{figure}

Given the proprietary nature of Llama's training data, we utilize RedPajama~\cite{together2023redpajama}, a 1.4 trillion token corpus designed to closely replicate Llama's dataset, as a stand-in. We use the Llama tokenizer for tokenization and examine occurrences of uni-grams across three categories: subparts ($N=206$), parts ($N=40$), and objects ($N=11$). We leverage the $\infty$-gram~\cite{liu2024infini} API for counting these occurrences within the RedPajama dataset. We observe decline in average uni-gram occurrence frequency when increasing granularity (\eg, part to subpart). This trend is depicted in Fig.~\ref{fig: token_dist}a. Further analysis of parts and subparts n-grams (Fig.~\ref{fig: token_dist}b) reveals that subpart n-grams (\eg, `eyes of the quadruped') are significantly less frequent, with an average of 7 instances, compared to parts (\eg, `head of the quadruped'), which average 75 instances. 

\section{ViP-Llava Adversarial Prompting}
We conducted two different adversarial prompting experiments for ViP-Llava 13B to better understand its near-perfect performance on interactive understanding. First, we conducted an adversarial experiment where we prompted the model the same way as the original experiment but randomly swapped out the object category for a different one among our set of object super-categories. As a consequence, the answer to the question, \texttt{"Is there an <object> in this <region>?"} is always `no.' We observe for this experiment that within object categories, mean accuracy decreases to 73.26\% (-25.07pp) for objects, 98.08\% (-1.82pp) for parts, and 96.62\% (-2.73pp) for subparts. These findings suggest that the inclusion of granular phrases (\eg, cheek) can help calibrate a model's confidence and reduce hallucinations, potentially due to the intrinsic associations it may make (\eg, recognizing that a bicycle does not have a cheek). Second, we prompted the model with the negation of the original prompt, \texttt{"Is there not an <object> in this <region>?"}, in which the answer is always `no'. Overall, we observe large decrease in performance with a mean accuracy of 28.36\% (13.65\% specific) for objects, 3.23\% (7.33\% specific) for parts, and 4.74\%  (4.17\% specific) for subparts. This big difference in performance from the results in the main paper reinforces findings from prior work that models struggle with negation~\cite{berglund2023reversal}. Moreover, these results highlight the importance of adversarial prompting and red-teaming foundation models to probe their biases (\eg, through tools like VLSlice~\cite{slyman2023vlslice}), such as a predisposition to answering yes to content that is not present within an image.

\section{Qualitative Results}
\subsection{Foundation Model Results}
\begin{figure}
    \centering
    \includegraphics[width=\textwidth]{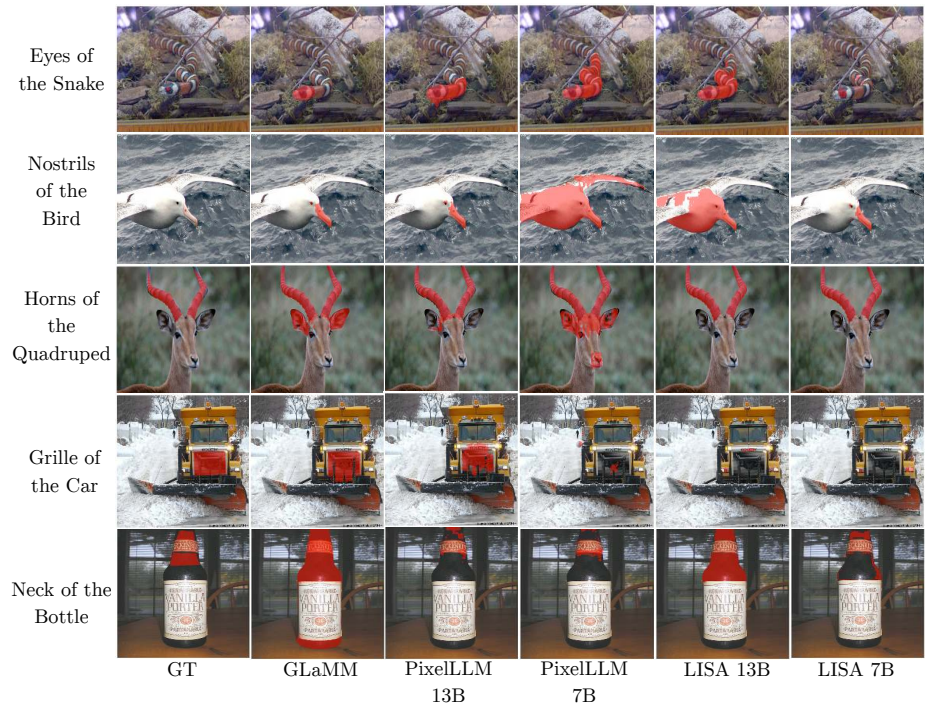}
    \caption{Qualitative results of models producing segmentation predictions, shown in red. Each row represents a different subpart. Columns display, from left to right: ground truth segmentations, followed by predictions from each method. For visualization purposes, all images are resized to square aspect ratios.}
    \label{fig: seg_qual}
\end{figure}

\begin{figure}
    \centering
    \includegraphics[width=\textwidth]{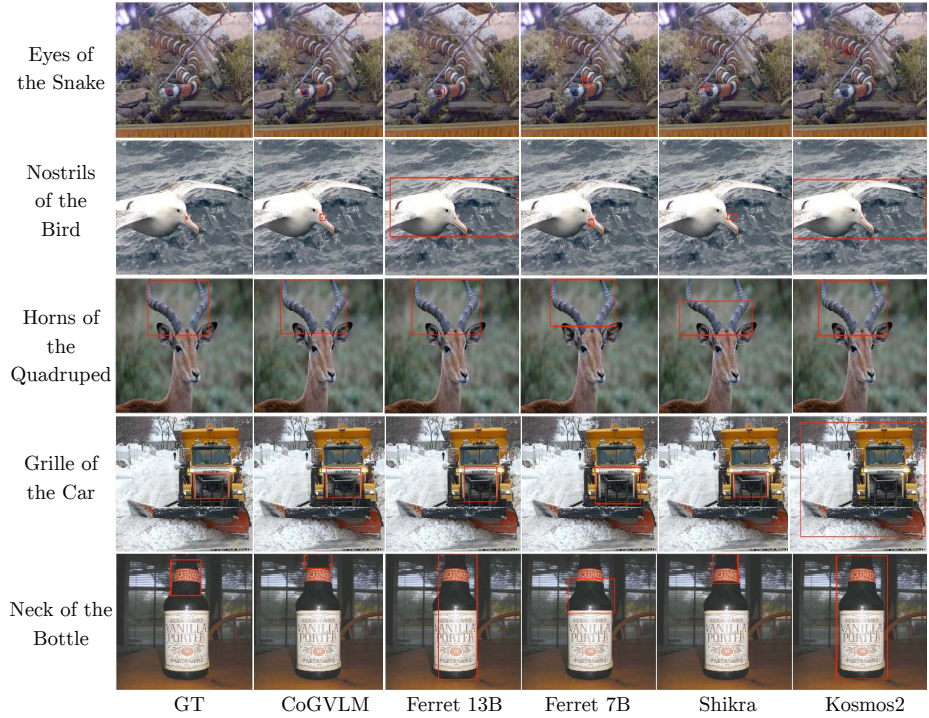}
    \caption{Qualitative results of models producing bounding box predictions, shown in red. Each row represents a different subpart. Columns display, from left to right: ground truth bounding boxes, followed by predictions from each method. For visualization purposes, all images are resized to square aspect ratios.}
    \label{fig: bbox_qual}
\end{figure}

Qualitative results for open-vocabulary object localization models are shown for 5 diverse examples in Fig.~\ref{fig: seg_qual} (segmentation) and Fig.~\ref{fig: bbox_qual} (object detection). We show examples for tiny subparts (\ie, eyes of the snake, nostrils of the bird) and large subparts (\ie, horns of the quadruped, neck of the bottle, grille of the car). 

For models capable of segmentation (LISA 7/13B, PixelLLM 7/13B, GLaMM), varied results are observed across examples. LISA 7B localizes snake eyes most accurately, while others locate the entire head or body portions. No model precisely segments bird nostrils, with the closest attempts segmenting the beak. Only LISA variants perfectly segment antelope horns without additional regions. For the grille, GLaMM provides the best segmentation, albeit with missing cruft. Regarding the bottleneck, LISA 13B achieves near-precise segmentation (aside from the inclusion of the shoulder), whereas other models either segment partial regions (LISA 7B, PixelLLM 7/13B) or all regions except the main label on the bottle (GLaMM).

For models that produce bounding boxes (CoGVLM, Ferret 7/13B, Shikra, Kosmos2), relatively consistent results are observed across examples. CoGVLM precisely locates tiny subparts (eye, nostril), while others produce shifted or object-encompassing bounding boxes. For larger regions (grille, horns, neck), all models except Shikra correctly localize antelope horns, and all except Kosmos2 accurately locate the car grille. Conversely, Shikra provides the closest bounding box for the bottleneck, with other models only capturing partial or complete bottle regions.

Overall, these results support our quantitative findings, with all models generally performing poorly on subpart localization. Overall, CoGVLM produces the best results, aligning with its superior quantitative performance.

\subsection{HIPIE Results}
\begin{figure}
    \centering
    \includegraphics[width=\textwidth]{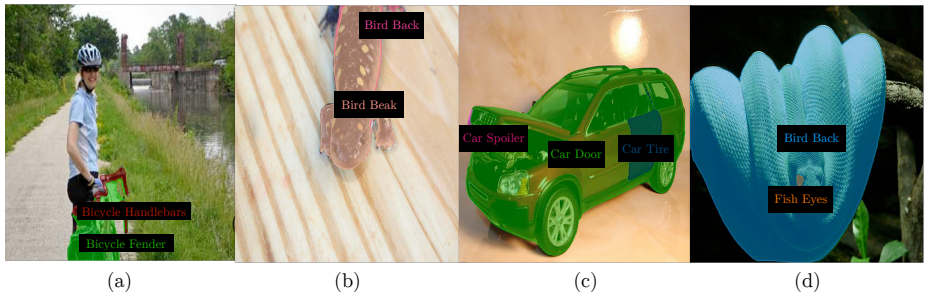}
    \caption{Qualitative results from HIPIE, with each panel showing all predicted segmentations with their corresponding label classification depicted in the same color. For visualization purposes, all images are resized to square aspect ratios.}
    \label{fig: hipie}
\end{figure}

Qualitative results for predicted subparts by HIPIE are shown in Fig.~\ref{fig: hipie}. (a) Shows a partially correct segmentation of bicycle handlebars, with incorrect labeling of the rest as ``fender''. (b) Demonstrates an out-of-distribution sample with incorrect labeling. (c) Exhibits a small number of correct class labels (``door'' and ``tire'') but with inaccurate segmentations. (d) Displays a semantically incoherent combination of ``bird back'' and ``fish eyes'', highlighting the need for holistic evaluation of granular segmentations (\eg, our proposed consistency scores). Overall, these poor results corroborate the quantitative findings reported in the main paper.

